# Deep neural network-based classification model for Sentiment Analysis

Donghang Pan
*College of Computer Science and Technology*
Wuhan University of Technology
Wuhan, China
pdonghang@126.com

Jingling Yuan
*College of Computer Science and Technology*
Wuhan University of Technology
Wuhan, China
yuanjingling@126.com

Lin Li
*College of Computer Science and Technology*
Wuhan University of Technology
Wuhan, China
cathylilin@whut.edu.cn

Deming Sheng
*College of Computer Science and Technology*
Wuhan University of Technology
Wuhan, China
2897102273@qq.com

*Abstract*—The growing prosperity of social networks has brought great challenges to the sentimental tendency mining of users. As more and more researchers pay attention to the sentimental tendency of online users, rich research results have been obtained based on the sentiment classification of explicit texts. However, research on the implicit sentiment of users is still in its infancy. Aiming at the difficulty of implicit sentiment classification, a research on implicit sentiment classification model based on deep neural network is carried out. Classification models based on DNN, LSTM, Bi-LSTM and CNN were established to judge the tendency of the user's implicit sentiment text. Based on the Bi-LSTM model, the classification model of word-level attention mechanism is studied. The experimental results on the public dataset show that the established LSTM series classification model and CNN classification model can achieve good sentiment classification effect, and the classification effect is significantly better than the DNN model. The Bi-LSTM based attention mechanism classification model obtained the optimal R value in the positive category identification.

*Keywords—Natural language processing, Chinese implicit sentiment, deep neural network, attention mechanism*

## I. INTRODUCTION

In multi-source social media data mining, research based on machine learning and deep learning has become the main method of pattern mining. For example, clustering can identify similar patterns[1-3], deep neural networks can provide end-to-end solutions[4-8], and feature embedding can provide problematic migration solutions[9].With the wide application of social media, rich text information has accumulated on the Internet. These subjective texts contain rich behavioral information of the members of society. Sentiment analysis of these texts can provide a deeper understanding of people's preferences for things, public events, and the users life pattern. It has far-reaching effects in public opinion analysis[10], resource recommendation[11], and law mining.

As a basic research task of sentiment analysis, user sentiment classification has made rich research progress in various aspects. According to the task objectives, sentiment classification can be divided into common polarity classification and emotion classification. According to the presence of sentimental words, the sentiment classification task can be divided into two types, including explicit sentiment classification and implicit sentiment classification. In social media based on micro-blog, users often display sentimental words in their expressions. For example, the food is delicious, where the word delicious implies a positive polarity. In actual research, social media users also use factual statements to express their sentimental attitudes. For example, your company's sales for one year can't keep up with us for a month. The user uses the facts to express his negative polarity. Unlike explicit expressions, implicit expressions do not contain sentimental words. Due to the large proportion of explicit text expression, the task of sentiment classification has achieved rich research results in the field of social media mining. In contrast, the study of implicit sentiment classification is still in its infancy. Implicit sentiment classification is an important part of sentiment analysis. Its research results will help to improve the performance of online text sentiment analysis more comprehensively and accurately. It can play a positive role in text representation learning, natural language understanding, user modeling, knowledge embedding and other research. It can also further promote the application of related fields and the rapid development of industry.

Compared with the task of explicit sentiment classification, the task of Chinese implicit sentiment classification is more difficult. From the linguistic level of sentimental expression, there is no change in the morphology of Chinese words[12]. It is more difficult to acquire Chinese semantic features. In terms of the words used, the explicit text contains sentimental words, which can help to identify the tendency of sentences to a large extent. Most implicit subjective texts use relatively neutral statements and do not contain sentimental words, which makes it difficult to distinguish the tendentiousness of sentences. Word vector representation technology based on neural language model can learn effective feature representation from large-scale corpus, and has made important progress in many natural language processing tasks. For example, the word2vec [13]model can capture similarities and differences between words. On the basis of the literature, the paper [14]proposes a sentence-level text feature representation method. Using word embedding technology to extract text features and training appropriate classifiers to classify sentence polarity has become an important technical means of sentiment classification tasks. Based on this idea, this paper uses the word2vec model to obtain the characteristics of sentences. Deep neural network-based classifier were established to conduct a biased classification study on Chinese implicitly expressed texts. The main research contents of this paper are as follows:

- Based on the deep neural network, the task of Chinese implicit sentimental polarity classification is studied. The deep neural networks used include convolutional neural network(CNN), deep fully connected neural network(DNN) and long short-term memory(LSTM). Since bidirectional LSTM(Bi-LSTM) has better effect in context semantic acquisition, a classification model based on Bi-LSTM is introduced in Chinese implicit sentiment classification. It is found that although the implicitly expressed text lacks the explicit sentiment words and is hard to classify the polarity, deep neural network structures proposed can achieve good classification results.

- Based on the idea of word-level attention mechanism, a classification model based on bidirectional LSTM (Bi-LSTM)structure is established. The results show that the attention mechanism implemented in this paper has achieved the optimal F1 value in the negative category identification. But this model has not achieved a significant performance improvement effect because of the missing of the sentiment words.

The second section of this paper introduces the related work of word embedding technology and deep learning model. The third section describes the various deep neural network models and attention-based classification models proposed. The fourth section introduces the experimentally relevant data sets and operational procedures, as well as the experimental results. Finally, the research work is summarized and forecasted.

## II. RELATED WORK

The sentiment classification task mainly consists of two parts, which are text feature representation and classification model selection. Related studies have shown that the length and quantity of text used for sentiment analysis in social media is extremely uneven[15]. The traditional word bag method has the problem of sparse data and dimensional explosion when performing text feature representation. Word embedding technology can handle this problem well. Compared with other languages, Chinese has great differences in semantic rules and social characteristics. Based on the word2vec technology, the literature introduces the n-gram feature into the context, uses the co-occurrence statistics of word-word and word-character to learn the word vector, and provides a method for judging the analogy of words[16]. The literature adds a variety of grammatical feature information when training the word vector, and uses a multi-domain corpus. Based on its research, this paper makes a feature representation of Chinese implicit sentimental text.

In the explicit text sentiment classification study, the classification method based on the sentiment dictionary weights the scores of the sentiment words to determine the overall polarity of the sentences. The literature analyzes the polarity of the evaluation words, weights and sums the polarities, and realizes the sentiment classification at the sentence level[17]. The machine learning-based classification method achieves sentiment classification of different levels of text by finding suitable features. The literature establishes Naive Bayes classification method and support vector machine method. The experimental results on a large number of data sets show that the Naive Bayes method achieves high precision[18]. The literature adds the word order relationship based on the existing LDA model, and uses the word sequence flow representation method to express the meaning of the word more accurately[19]. A summary of the classification method based on machine learning found that the method has the problem that the features are not easy to extract. Due to the lack of sentimental words in Chinese implicit expression , it is impossible to determine which type of features are more suitable for implicit sentiment classification. Therefore, the sentiment classification method based on machine learning is not suitable for the field of Chinese implicit sentiment classification.

The sentiment classification method based on deep neural network provides an end-to-end problem solution, and can abstract the representation and synthesis of word features, which has become the mainstream research method in NLP field. In order to analyze the sentiment of specific aspects[20], the paper proposes a specific aspects sentiment analysis method for multi-conflict convolutional neural networks, obtains deeper sentimental feature information, and effectively recognizes the sentiment polarity of different aspects. In this paper, a Chinese microblog sentiment analysis model based on multi-channel convolutional neural network is proposed[21]. This model can effectively obtain the sentimental feature information of input sentences, and can effectively represent the importance of each word in the sentence. Experimental results show that this method has better performance than ordinary convolutional neural networks, convolutional neural networks combined with emotional information and traditional classifiers. In this paper, the word embedding technology is used to obtain the emoji of micro-blog, and the sentimental space matrix is constructed[22]. On this basis, a multi-channel convolutional neural network classifier is trained, and the best classification performance is obtained on the NLPCC microblog evaluation data set. The literature [23]uses a variety of deep neural networks to solve cross-domain sentiment analysis tasks. Experimental results show that deep neural networks can achieve better classification results. Most of the above-mentioned literature research content is explicit text. For example, some models can make sentimental words have a higher proportion in sentences through unique design. In the field of implicit sentiment classification tasks, related deep neural network research is lacking. Taking this as a starting point, the classification of basic implicit neural networks of CNN, DNN,LSTM and Bi-LSTM is carried out on Chinese implicit texts.

Due to the existence of the display of sentimental words, the attention mechanism is introduced in the classification model. By giving higher weights to sentimental words, sentence performance classification performance can be improved. In the basis of mind mechanism of attention, this paper implements a classification model of attention level at the word level[24]. This paper establishes a word-level attention mechanism classification model to realize the classification of implicit sentiment texts.

## III. MODEL DESCRIPTION

### A. Classification model framework

The classification framework based on deep neural network mainly includes three parts. From the bottom up, the underlying word embedding layer, the deep neural network layer, and the top softmax layer. The overall structure of the frame is shown in Figure 1.

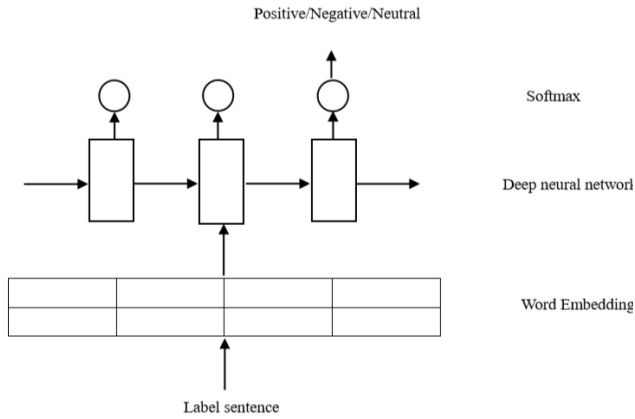

Fig. 1. Classification model frame.

The difference between different classification models is the network structure selected by the deep neural network layer. The underlying word embedding layer and the top-level softmax classification layer use the same structure. The underlying word embedding layer and the top-level softmax classification layer use the same structure. The word pre-training technique is used in the word embedding layer to represent sentence features. The word embedding technology has obtained rich research results in text feature representation. The rich vector feature representation can be obtained through the bottom pre-training model. In order to obtain the sentence context representation, the word embedding layer of the sentence is obtained based on the word2vec model. The Wikipedia Chinese word vector matrix provided in the literature was obtained[16]. The word embedding layer is constructed to obtain the mapping relationship between words and vectors in sentences. The relevant training parameters of the model are shown in Table 1.

TABLE I. WORD2VEC PARAMETERS

| Window size | Dynamic window | Sub-sampling | Low-Frequency Word | Iteration | Negative Sampling |
|---|---|---|---|---|---|
| 5 | Yes | 1e-5 | 10 | 5 | 5 |

Four different structures are used in the deep neural network layer, namely DNN, CNN, LSTM and Bi-LSTM. This layer is responsible for feature extraction and synthesis of the underlying word vector. In order to weaken the over-fitting phenomenon of the deep neural network, a dropout mechanism is added after constructing a layer[25]. The four network model structures are described in detail in the following subsections. The softmax activation function is used in the top level classification layer to obtain the probability of each category.

*B. LSTM and its variants*

In actual language expression, the relationship between words constitutes a complete statement. Different combinations of words can produce different semantics. The premise of convolutional neural networks and artificial neural networks is that elements are independent of each other. Inputs and outputs are independent. When the classification model is established by using the above two network structures, the relationship of word features is separated. In contrast, LSTM is a time recurrent neural network that assumes that the output at the current moment depends on the current input and past inputs. This model structure can acquire time-dependent features well. Valid feature information can be captured in long sequence text. LSTM is available in one-to-many, many-to-one and many-to-many formats and can be used in text categorization, sentiment analysis, and machine translation. Due to LSTM's unique feature extraction approach and outstanding performance in text sentiment analysis. This paper builds a basic LSTM sentiment classification model to classify Chinese implicit texts. The LSTM neural network layer structure is shown in Figure 1. From the image we can see that the LSTM neurons are related to each other. This is a one-way LSTM structure that only has a front-to-back effect.

Unlike LSTM, Bi-LSTM is a bidirectional LSTM structure. When LSTM acquires text features, it only establishes the influence information of the previous text onthe following text. Based on this, Bi-LSTM captures bidirectional semantic information. In certain mission scenarios, Bi-LSTM can achieve better performance. In order to study whether the Bi-LSTM structure is more suitable for Chinese implicit text classification, this classification model is established in this paper. The model structure is shown in Figure 2.

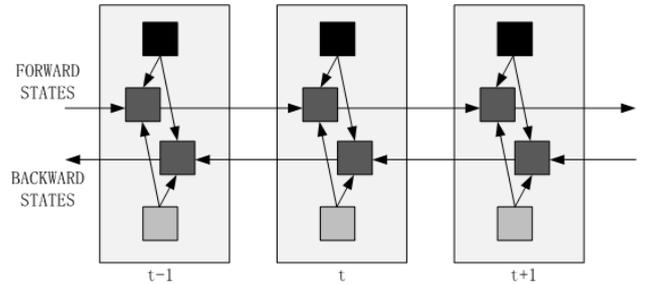

Fig. 2. Bi-LSTM structure.

*C. CNN and DNN*

Although CNN and artificial neural networks separate the complex relationships between elements, both can extract and synthesize features through complex network structures. CNN can capture local important information through convolution and pooling operations. Therefore it is also applied to related natural language processing tasks. Multi-layer fully connected neural networks(DNN) can be used to fit features through complex structures. The literature uses CNN to make effective attempts in the field of NLP[26]. A sentence is represented as a matrix by splicing the word vector. A set of convolutional layer filters and a maximum pooling layer of the pooling layer are used to obtain the feature vectors of the sentences, and finally the classification of the sentences is realized. Combine the literature ideas ,in this paper, a set of one-dimensional convolutional layer and maximum pooling layer are designed in the deep neural network layer to realize the extraction and synthesis of text features. In the feature output map, the first convolutional layer is padded with zero and the second convolutional layer is not padded with zero. The model structure is shown in Figure 3.

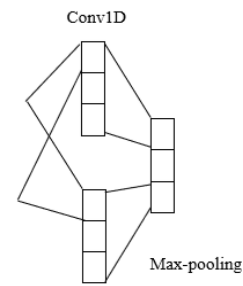

Fig. 3. CNN structure.

The ability to synthesize model features can be increased by constructing a multi-layer fully connected layer neural network. In the DNN classification model, a three-layer fully connected layer is designed. The model structure is shown in Figure 4.

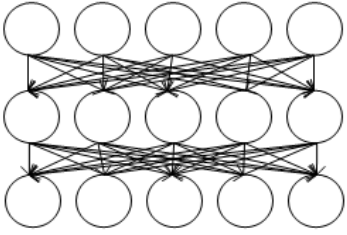

Fig. 4. DNN structure.

### D. Classification model based on sentence attention mechanism

Compared to the LSTM model, the LSTM uses the update gate structure instead of the forget gate and input gate in the LSTM. The reset gate is used to control the degree of ignoring the status information of the previous moment. This design allows the LSTM to maintain the LSTM effect while streamlining the network structure, resulting in less computation. The literature designs the attention of two mechanisms, including word level and sentence level[24]. Based on the idea of word-level attention mechanism proposed in the above literature, this paper implements the attention mechanism based on Bi-LSTM. The network structure diagram is shown in Figure 5.

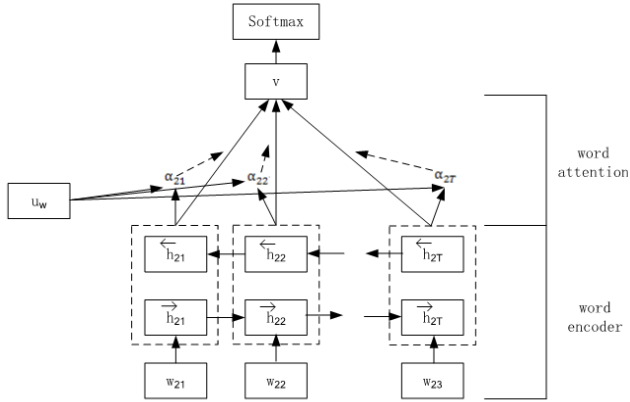

Fig. 5. Bi-LSTM based attention structure.

The appropriate weights are assigned to the input by the method of the below formula, and a fully connected layer is added to the output part to realize the synthesis of the features. The input features are weighted differently by assigning weights, and a fully connected layer is added to the output part to realize the synthesis of the features. The core formula is shown in formula (1)(2)(3), as in:

$$\mu_{it} = \tanh(W_w h_{it} + b_w) \quad (1)$$

$$\alpha_{it} = \frac{\exp(\mu_{it}^T \mu_w)}{\sum_t \exp(\mu_{it}^T \mu_w)} \quad (2)$$

$$S_i = \sum_t \alpha_{it} h_{it} \quad (3)$$

### E. Evaluation Index

The evaluation criteria used in the model evaluation include precision rate (P), recall rate (R) and F1 value. Since the classification category includes three, one is a positive example, and the other two are negative examples to calculate three evaluation indicators. The final calculation of the average value is the final evaluation standard of the model.

## IV. EXPERIMENT AND ANALYSIS

### A. Dataset description

This paper uses the Evaluation of Chinese Implicit Sentiment Analysis,(SMPECISA 2019) to conduct experiments. Since the test set has not yet been pub lished, the data set provided contains only the training data and the validation data. The training set is internally divided according to a ratio of 4:1 for the training process of the model. The officially provided validation set is used as a test set to evaluate the performance of each classification model. The data includes microblogs, travel websites, product forums and other fields. The main topics include Spring Festival Evening, Haze, National Examination, Tourism, Dragon Boat Festival and so on. The sentiment label contains a total of positive, negative, and neutral. The detailed quantity and related context information are shown in Table 2. Among them, the official training set and verification set contain 33 co-occurring sentences, and this article chooses to remove them in the test set.

TABLE II. DATASET ATTRIBUTE

| | Label sentence | Positive | Negative | Neutral |
|---|---|---|---|---|
| train | 14774 | 3828 | 3957 | 6989 |
| test | 5110 | 1222 | 1355 | 2533 |

### B. Experiment process

A preliminary rule identification is performed on the experimentally obtained data set, and the tag sentence and the context are respectively extracted. Since the text content comes from the network, the original data set contains a large number of network terms. In order to preserve the characteristics of the original text to the greatest extent, the preprocessing stage only removes the sentences co-occurring in the test data. After the sentence extraction is completed, the sentence is processed by the jieba word segmentation tool.

A set of word vectors incorporating the three features described above is employed. When this vector representation is used, words that do not appear in the corpus fill the vector with 0. The training set is divided into a training data and a validation data according to a ratio of 4:1, and the validation data is used to adjust the prediction effect of the classification model. The ratio of the training data to the test data is approximately 3:1. Three replicate experiments were performed on all models, and the average was used to evaluate the classification effect of each model. The model also includes some parameter settings. The unique settings are shown in the table below, and the rest of the parameters are default values.

TABLE III. MODEL PARAMETERS

| Parameters | Values |
|---|---|
| word vector dimension | 300 |
| epoch | 10 |
| dropout | 0.5 |
| DNN dimension | 128,64,32 |
| LSTM dimension | 64 |
| CNN filter | 300 |

*C. Results and analysis*

In the actual experimental results, after about 10 epoch, the model is basically well trained. To reduce over-fitting and under-fitting, the epoch has a value of 10. In the trained model, a classification model with numerical values tending to be stable is selected for testing.

In this paper, the datasets are classified by ternary polarity using five classification models. The categories included are positive, neutral, and negative. The experimental results are shown in Table 4. Implicit text is more difficult to classify in ternary sentiment than explicit sentiment classification.

TABLE IV. THREE-CATEGORY CLASSIFICATION RESULT

| Model | Index | Neutral | Positive | Negative |
|---|---|---|---|---|
| DNN | P | 81.22 | 51.38 | 65.27 |
| | R | 84.48 | 45.85 | 65.66 |
| | F1 | 82.76 | 48.27 | 64.86 |
| CNN | P | **84.37** | 60.02 | 72.37 |
| | R | 86.31 | **57.34** | 71.27 |
| | F1 | **85.29** | 58.51 | 71.37 |
| LSTM | P | 84.87 | 58.29 | 73.18 |
| | R | 84.51 | 61.02 | 70.43 |
| | F1 | 84.64 | **59.57** | 71.74 |
| Bi-LSTM | P | 83.22 | **61.46** | 70.76 |
| | R | 86.29 | 54.31 | **73.21** |
| | F1 | 84.71 | 57.63 | **71.95** |
| Bi-LSTM based attention | P | 81.65% | 59.76% | 73.89% |
| | R | **87.80%** | 57.01% | 65.98% |
| | F1 | 84.58% | 58.16% | 69.60% |

It can be seen from Table 4 that CNN, Bi-LSTM, LSTM and attention-based classification models are significantly superior to the DNN classification model in the three-category implicit sentiment classification task. The CNN classification model has better performance in neutral category identification. The LSTM classification model has a good F1 value in positive category identification. The established word-level attention mechanism classification model has better overall performance in negative category identification.

From the table, we can find that the classification model of the LSTM series can extract better correlations between features, and thus have achieved good performance. CNN's pooling operation can extract local important information, and thus achieves good results in classification results. DNN only has the ability to nonlinearly fit, and in the feature extraction of different lengths, it does not achieve a good classification effect. Because the LSTM series classification model itself has the ability to assign different weights to input features. In the implicit expression, there are no particularly important sentimental words, so the attention mechanism implemented in this paper has not achieved a significant performance improvement effect. This also shows that implicit expression classification is more difficult than explicit expression.

V. SUMMARY AND PROSPECT

This paper classifies Chinese implicit sentiment based on deep neural networks. The established LSTM, CNN and Bi-LSTM basic classification models have achieved better classification results than DNN. Based on Bi-LSTM, this paper also implements a classification model based on attention mechanism. The analysis of the two-category results and three-category experiments found that the LSTM series and the CNN classification model achieved good classification results due to the unique design of the feature extraction. Since the implicit sentimental text does not include the sentimental words, the classification model based on the attention mechanism has not been significantly improved. However, the implementation of the attention-based mechanism model has achieved the best results in the classification of some categories. The experimental results also show that implicit sentimental text categorization is more difficult than explicit sentimental text categorization tasks.

REFERENCES

[1] Y. Wang, X. Lin, L. Wu, et al, Robust subspace clustering for multi-view data by exploiting correlation consensus IEEE Transactions on Image Processing, 24(11):3939-3949, 2015.

[2] Y. Wang, L. Wu, X. Lin, J. Gao. Multiview Spectral Clustering via Structured Low-Rank Matrix Factorization. IEEE Transactions on Neural Networks and Learning Systems 29 (10), 4833-4843, 2018.

[3] Y. Wang, W. Zhang, L. Wu et al., Iterative Views Agreement: An Iterative Low-Rank based Structured Optimization Method to Multi-View Spectral Clustering. IJCAI 2016.

[4] Y. Wang, X. Lin, L. Wu, W. Zhang. Effective Multi-Query Expansions: Collaborative Deep Networks for Robust Landmark Retrieval. IEEE Transactions on Image Processing 26 (3), 1393-1404, 2017.

[5] L. Wu, Y. Wang, X. Li, J. Gao. Deep Attention-based Spatially Recursive Networks for Fine-Grained Visual Recognition. IEEE Transactions on Cybernetics 49 (5), 1791-1802, 2019.

[6] L Wu, Y Wang, X Li, J Gao. What-and-Where to Match: Deep Spatially Multiplicative Integration Networks for Person Re-identification Pattern Recognition 76, 727-738, 2018.

[7] L. Wu, Y. Wang, L. Shao. Cycle-Consistent Deep Generative Hashing for Cross-Modal Retrieval. IEEE Transactions on Image Processing 28 (4), 1602-1612, 2019.

[8] L Wu, Y Wang, J Gao, X Li. Where-and-When to Look: Deep Siamese Attention Networks for Video-based Person Re-identificationIEEE Transactions on Multimedia 21 (6), 1412-1424, 2019.

[9] L Wu, Y Wang, J Gao, X Li. Deep Adaptive Feature Embedding with Local Sample Distributions for Person Re-identification, Pattern Recognition 73, 275-288, 2018.

[10] Huang FL, Yu G, Zhang JL, Li CX, Yuan CA, Lu JL. Mining topic sentiment in micro-blogging based on microblogger social relation. Journal of Software, 2017,28(3):694-707.

[11] Huang FL, Feng S, Wang DL and Yu G.Mining topics sentiment in Microblogging based multi-feature fusion.[J]Chinese Journal of computer, 2017, 40(4): 872-888.

[12] Yuanyuan Qiu, Hongzheng Li, Shen Li, Yingdi Jiang, Renfen Hu, Lijiao Yang. Revisiting Correlations between Intrinsic and Extrinsic Evaluations of Word Embeddings. Chinese Computational Linguistics and Natural Language Processing Based on Naturally Annotated Big Data. Springer, Cham, 2018. 209-221.

[13] Mikolov T, Chen K, Corrado G, et al. Efficient Estimation of Word Representations in Vector Space[J]. Computer Science(CS), 2013.

[14] Le Q, Mikolov T. Distributed representations of sentences and documents[C]//International conference on machine learning. 2014: 1188-1196.

[15] Zhang L, Qian GQ, Fan WG, Hua K, Zhang L. Sentiment analysis based on light reviews[J].Journal of Software, 2014,25(12):2790-2807 .

[16] Li S, Zhao Z, Hu R, et al. Analogical Reasoning on Chinese Morphological and Semantic Relations[J]. 2018.

[17] Yu H, Hatzivassiloglou V. Towards answering opinion questions: Separating facts from opinions and identifying the polarity of opinion sentences[C]//Proceedings of the 2003 conference on Empirical methods in natural language processing. Association for Computational Linguistics(ACL), 2003: 129-136.

[18] Wawre S V, Deshmukh S N. Sentiment classification using machine learning techniques[J]. International Journal of Science and Research (IJSR), 2016, 5(4): 819-821.

[19] Griffiths T L, Steyvers M, Tenenbaum J B. Topics in semantic representation[J]. Psychological review(PR), 2007, 114(2): 211.

[20] Liang B,Liu Q,Xu J.et al.Aspect-based sentiment analysis based on multi-attention CNN[J].Journal of computer research and development,2017,54(8):1724-1735.

[21] Chen K,Liang B,Ke WD,et al.Chinese micro-blog sentiment analysis based on muti-channels convolutional neural networks[J]. Journal of computer research and development,2018,55(5):945-957.

[22] He YX,Sun ST,Niu FF and Li F.A deep-learning model enhanced with emotion semantics for microblog sentiment analysis[J]. Chinese journal of computers,2017(4).

[23] Yu J, Jiang J. Learning sentence embeddings with auxiliary tasks for cross-domain sentiment classification[C]//Proceedings of the 2016 conference on empirical methods in natural language processing(NLP). 2016: 236-246.

[24] Yang Z, Yang D, Dyer C, et al. Hierarchical attention networks for document classification[C]//Proceedings of the 2016 Conference of the North American Chapter of the Association for Computational Linguistics: Human Language Technologies. 2016: 1480-1489.

[25] Srivastava N, Hinton G, Krizhevsky A, et al. Dropout: a simple way to prevent neural networks from overfitting[J]. The Journal of Machine Learning Research, 2014, 15(1): 1929-1958.

[26] Kim Y. Convolutional neural networks for sentence classification[J]. arXiv preprint arXiv:1408.5882, 2014.